\documentclass{article}

\usepackage[normalem]{ulem}

\usepackage{microtype}
\usepackage{graphicx}
\usepackage{subfigure}
\usepackage{booktabs} 
\usepackage[export]{adjustbox}
\usepackage{csquotes}
\usepackage{framed}

\usepackage{hyperref}

\usepackage[accepted]{icml2025}

\usepackage{amsmath}
\usepackage{amssymb}
\usepackage{mathtools}
\usepackage{amsthm}
\usepackage{comment}
\usepackage[abs]{overpic}

\usepackage{pifont}
\usepackage[capitalize,noabbrev]{cleveref}

\theoremstyle{plain}

\theoremstyle{definition}

\theoremstyle{remark}

\usepackage[textsize=tiny]{todonotes}

\icmltitlerunning{Guidance Free Image Editing via Explicit Conditioning}

\begin{document}

\twocolumn[
\icmltitle{Guidance Free Image Editing via Explicit Conditioning}



\icmlsetsymbol{equal}{*}

\begin{icmlauthorlist}
\icmlauthor{Mehdi Noroozi}{}
\icmlauthor{Alberto Gil Ramos}{}
\icmlauthor{Luca Morreale}{}
\icmlauthor{Ruchika Chavhan}{}
\icmlauthor{Malcolm Chadwick}{}
\icmlauthor{Abhinav Mehrotra}{}
\icmlauthor{Sourav Bhattacharya}{}

\icmlauthor{}{}

\icmlauthor{Samsung AI Cambridge}{}

\href{m.noroozi@samsung.com}{m.noroozi@samsung.com}
\end{icmlauthorlist}

\icmlcorrespondingauthor{Mehdi Noroozi}{m.noroozi@samsung.com}


\vskip 0.3in
]




\newcommand{\sourav}[1]{\textcolor{red}{#1}}
\newcommand{\luca}[1]{\textcolor{blue}{#1}}
\newcommand{\Luca}[1]{\textcolor{blue}{Luca: #1}}

\newcommand{\mehdi}[1]{\textcolor{green}{Mehdi: #1}}

\newcommand{\eg}{e.g.}
\newcommand{\ie}{i.e.}

\newcommand{\N}{\mathcal{N}}
\newcommand{\x}{\mathbf{x}}
\label{sec:abstract}
\begin{abstract}

Current sampling mechanisms for conditional diffusion models rely mainly on Classifier Free Guidance (CFG) to generate high-quality images. 
However, CFG requires several denoising passes in each time step, \eg, up to three passes in image editing tasks,  resulting in excessive computational costs. 
This paper introduces a novel conditioning technique to ease the computational burden of the well-established guidance techniques, thereby significantly improving the inference time of diffusion models. We present Explicit Conditioning (EC) of the noise distribution on the input modalities to achieve this. Intuitively, we model the noise to guide the conditional diffusion model during the diffusion process. We present evaluations on image editing tasks and demonstrate that EC outperforms CFG in generating diverse high-quality images with significantly reduced computations.
\end{abstract}
\label{sec:intro}
\section{Introduction}
\begin{figure*}[ht!]
\begin{minipage}{0.3333\textwidth}
\includegraphics[width=\textwidth]{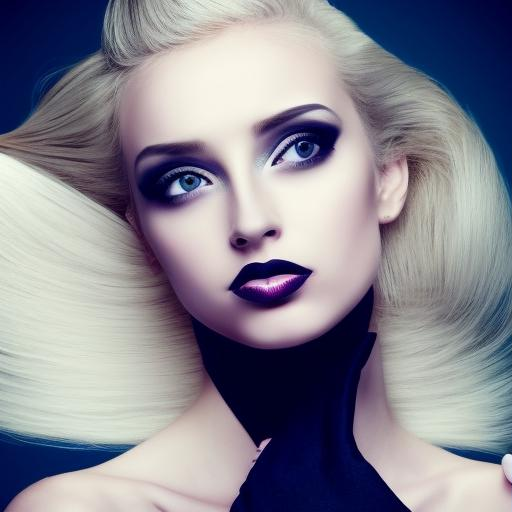}
\end{minipage}
\begin{minipage}{0.3333\textwidth}
\includegraphics[width=\textwidth]{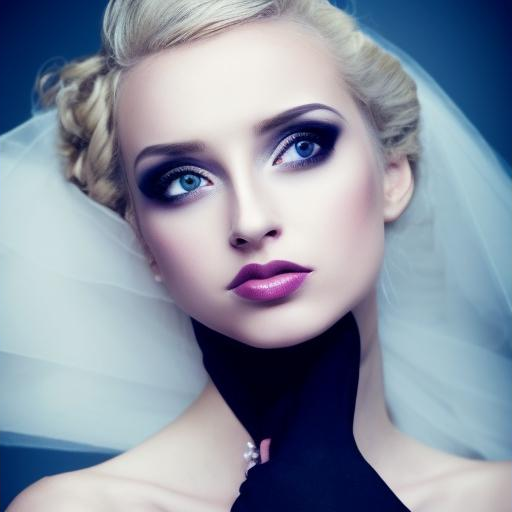}
\end{minipage}
\begin{minipage}{0.3333\textwidth}
\includegraphics[width=\textwidth]{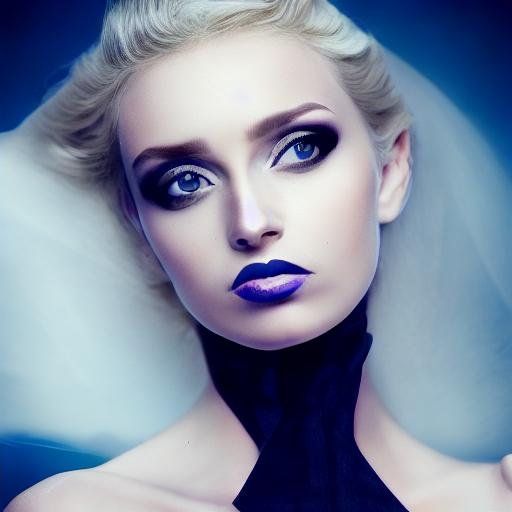}
\end{minipage}

\begin{minipage}{0.333\textwidth}
\centering
(a) context
\end{minipage}
\begin{minipage}{0.333\textwidth}
\centering
(b) target
\end{minipage}
\begin{minipage}{0.333\textwidth}
\centering
{(c) EC $\times 1$ pass}
\end{minipage}

\begin{minipage}{0.3333\textwidth}
\includegraphics[width=\textwidth]{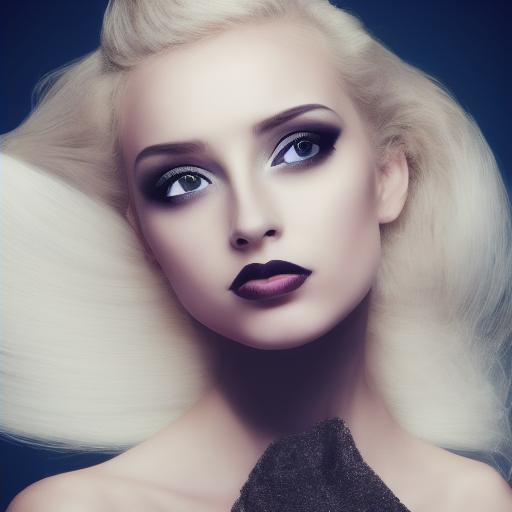}
\end{minipage}
\begin{minipage}{0.3333\textwidth}
\includegraphics[width=\textwidth]{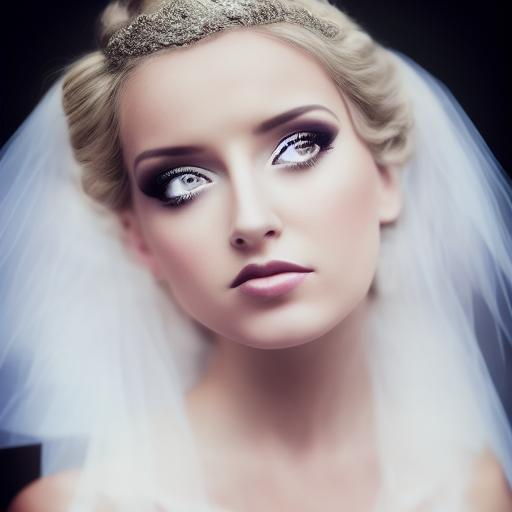}
\end{minipage}
\begin{minipage}{0.3333\textwidth}
\includegraphics[width=\textwidth]{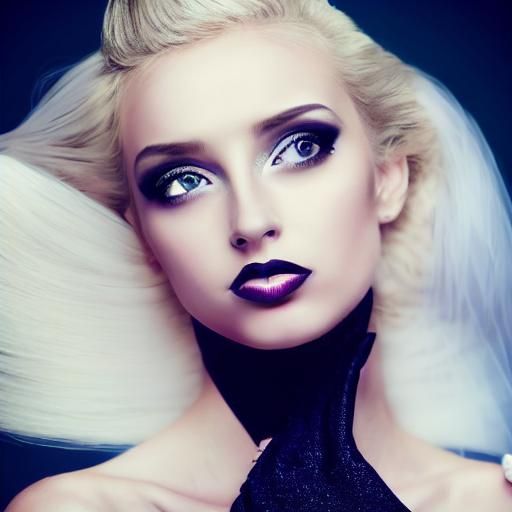}
\end{minipage}

\begin{minipage}{0.333\textwidth}
\centering
{(d) CFG $\times 1$ pass}
\end{minipage}
\begin{minipage}{0.333\textwidth}
\centering
{(e) CFG $\times 2$ passes} 
\end{minipage}
\begin{minipage}{0.333\textwidth}
\centering
{(f) CFG $\times 3$ passes}
\end{minipage}

\caption{Image editing performance for the context image in (a) with the instruction prompt: "Make her a bride". CFG (Eq.~\ref{eq:cfg implicit}) are shown with $\times 1$ pass ($s_I=1.0, s_P=1.0$) in (d),  $\times 2$ passes ($s_I=1.0, s_P=7.5$) in (e), and $\times 3$ passes ($s_I=1.6, s_P=7.5$) in (f). Our proposed explicit conditioning result with a single pass is shown in (c).}
\label{fig:teaser}
\end{figure*}
Diffusion models~\cite{Karras_2022,Karras_2024} have made great progress in image generation abilities, demonstrating impressive results for conditional generation.
Currently, all state-of-the-art (SOTA) conditional diffusion systems leverage Classifier Free Guidance (CFG)~\cite{CFG,Zheng_2023} to yield diverse high-quality conditional image generation, at the cost of the increased number of the denoising passes of the learned neural network models. Specifically, the image editing task is conditioned on a given image and an instruction prompt. Applying CFG for image editing requires three denoising passes to yield plausible results, see Fig.~\ref{fig:teaser} (d), (e), and (f). 

Informally, the formulation of CFG is based on two premises~\cite{CFG,Zheng_2023,brooks2022instructpix2pix}. First, there exists conditional information that is not natively supported by the underlying generative modeling. Second, the score, \ie, the gradient of the log of the density of conditional generative modeling may be aided by the score of unconditional generative modeling. More formally, CFG can be seen as predictor-corrector~\cite{Bradley_2024}. CFG boosts conditional image generation significantly across diffusion and flow~\cite{Lipman_2023,Tong_2024} models alike. This has led to a surge of research on different ways to combine not just conditional and unconditional models but more generally two or more models~\cite{Sadat_2024}.

The main practical drawback of CFG lies in the need for multiple conditional and unconditional denoising passes, resulting in additional computation costs. To alleviate the computational cost, a student model is distilled to mimic in one pass the original teacher sampling mechanism requiring multiple passes. CFG distillation, however, imposes a cumbersome extra training stage and is prone to lag behind the teacher's performance~\cite{meng23}. Moreover, the conditional and unconditional denoising passes need to be combined via a scaling factor where the optimal value is sample-dependent and requires a few trials by the user.

In this paper, we propose Explicit Conditioning (EC) for image editing. Rather than working on another candidate replacement for CFG or a distillation mechanism to cease the extra computational cost, we investigate generative modeling that natively supports conditional information. Instead of diffusing the image distribution into a noise distribution, and adding the $\mathrm{condition}$ as an additional input, we consider diffusing the data distribution into a purposefully designed distribution made of noise and $\mathrm{condition}$.
 Explicit conditioning alleviates the computational burden of CFG: the proposed generative framework natively incorporates conditional information, avoiding extra denoising passes during the sampling.

Fig.~\ref{fig:teaser} demonstrates that by explicitly conditioning the diffusion model, we diminish the need for guidance during the sampling. As it is shown in Fig.~\ref{fig:teaser} (d), (e) and (f), the current conditioning approach for image editing~\cite{brooks2022instructpix2pix} requires multiple denoising passes to achieve plausible results. By explicitly conditioning both context input and the instruction prompt,  the image editing framework using our EC approach outperforms CFG-based sampling using only $\times 1$ pass, being not only $\times 3$ faster but also improving the performance, Fig.~\ref{fig:teaser} (f). We support our claim via extensive quantitative and qualitative evaluations.

In the reminder of the manuscript, we first review concepts of diffusion models (Section~\ref{sec:method}). We leverage those principles to introduce Explicit Conditioning as a novel guidance-free image editing framework (Section~\ref{sec:explicit_conditioning}). Then, theoretically justify the effectiveness of explicit conditioning. Finally, we show quantitatively and qualitatively that our proposed explicit conditioning outperforms CFG while being $\times 3$ faster (Section \ref{sec:experiments}).

\section{Related Work}
\label{sec:related_works}
\paragraph{Conditional generative models.}
Image generation is dominated by diffusion and flow models which yield state-of-the-art (SOTA) diverse high-quality image generation, far surpassing GAN~\cite{Huang_2025, CGANs, SCGANs} alternatives. There has been a surge in the development of diffusions and flows.
\if
Hereafter we highlight SOTA methods for unconditional and conditional image generation~\ref{subsection: Latent and pixel based image generation}--\ref{subsection: Unifying diffusions and flows}. We emphasize key works on Classifier Free Guidance (CFG) and CFG-like techniques~\ref{subsection: CFG and selected alternatives}, distillation for reduced denoising network evaluations per step~\ref{subsection: Distillation}, conditioning via distributions rather than inputs~\ref{subsection: Conditioning via distributions rather than inputs} and image editing~\ref{subsection: Image editing}.
\fi

At a high level, diffusions and flows transform distributions into distributions. For example, unconditional image generation learns to transform the underlying image data distribution into a multivariate normal distribution. A key design choice, in this paradigm, is whether the transformation occurs in the original pixel space~\cite{Hoogeboom_2024,NVIDIA_2024} or in a latent space~\cite{Rombach_2022,Esser_2024}. SOTA approaches are dominated by the latter. 

Another key design criteria is whether the transformation from data into noise is done in a discrete number of time steps, or in a continuous range of time steps. Diffusion models owe their success to rapid progress on the discrete time setting, formally modeled by hierarchical VAEs~\cite{Luo_2022}. Notable examples include Denoising Diffusion Probabilistic Models (DDPM)~\cite{ddpm} and Denoising Diffusion Implicit Models (DDIM)~\cite{Song_2020}.

An important practical observation early on was that although diffusion models needed to be trained with a high number of discrete time steps, they could be sampled with a low number of time steps. Another key practical observation of diffusion models was that their training and sampling very much depended on the logarithm of the Signal-to-Noise Ratio (SNR) which was rigidly tied to the predefined discrete time steps. This let to the development of continuous time diffusion models, seen as the limit of hierarchical VAEs or as Stochastic Differential Equations (SDEs)~\cite{Song_20,Song_2021,Karras_2022,Karras_2024}. Despite the success of diffusion models, faster training and more diverse and higher quality sampling was shown to depend on selecting data, noise or velocity parametrizations~\cite{Salimans_2022}. Conditional Flow Matching (CFM)~\cite{Lipman_2023,Tong_2024,Lipman_2024,Esser_2024} provides an alternative natural parametrization for diffusion, namely via optimal transport, that yields SOTA models. More broadly, general diffusion and flow models can be recast in a unified framework, viewed under Evidence Lower BOund (ELBO) objectives~\cite{Kingma_Gao_2023}.

\paragraph{CFG and selected alternatives}
\label{subsection: CFG and selected alternatives}
Classifier Free Guidance (CFG)~\cite{CFG,Zheng_2023} is a key pillar of SOTA diffusion and flow systems, defining a conditional framework to image generation. 
Great focus is on its theoretical underpinning~\cite{Bradley_2024} and candidate alternatives such as Independent Condition Guidance (ICG) and Time Step Guidance (TSG)\cite{Sadat_2024}. Our work provides a new avenue of research for conditional image generation with a new Explicit Conditioning (EC) mechanism that is on par with CFG in terms of sample diversity and quality.

\paragraph{Single pass distillation. }
\label{subsection: Distillation}
Previous work~\cite{meng23} can distill CFG from three passes to one pass, sensibly reducing latency, while matching CFG on diversity and quality generation. More recently~\cite{Ahn2024} proposed to extend the distillation-based line of methods to learn a mapping from normal Gaussian to a guidance-free Gaussian distribution. Their work shares similarities with us in terms of exploring the sampling distribution, however, they still rely on guidance-dependent teachers to learn such a mapping.  Our approach achieves the same goal without the painstaking of distillation. 

\paragraph{Image editing.}
The most notable example of diffusion based image editing is~\cite{brooks2022instructpix2pix}, in which unedited image is transformed into edited image by passing the editing prompt to the denoising network, nobably by simple concatenation along the channel dimension on the first layer of Stable Diffusion (SD)~\cite{Rombach_2022}.

\paragraph{Editing via distributions rather than inputs.}
While the vast majority of diffusion and flow models transform image into noise through some process, there are exceptions that transform between more general data distributions. Notable instances include class transformations~\cite{Zhou_2023,McAllister_2024} and super resolution~\cite{Mei_2032}. No prior work transforms image and text based distributions into image distributions directly without text conditioning as an additional input as EC does. Although we focus our numerical experiments on image editing tasks, EC is applicable to a much wider class of conditional generation problems.

\label{sec:method}

\section{Method}
In this section, we discuss our proposed explicit conditioning (EC) in details. We first review diffusion models and classifier-free guidance (CFG) and discuss how EC is independent of guidance.

\subsection{Preliminaries}

The foundation of diffusion and flow image generation is in the unconditional setting, \ie, generate an image $x$ from a latent $\epsilon$ through a process
\begin{equation}
    z_t \coloneqq \alpha_tx + \sigma_t\epsilon,
    \label{eq: alpha x + sigma epsilon}
\end{equation}
with density
\begin{equation*}
    p(z_t) \coloneqq p(z_t|\epsilon)p(\epsilon).
    \label{eq: unconditional}
\end{equation*}
The conditional setting is based on combining the aforementioned density with
\begin{equation*}
    p(z_t|c) \coloneqq p(z_t|\epsilon, c)p(\epsilon),
    \label{eq: conditional}
\end{equation*}
in a gamma-powered distribution fashion
\begin{equation}
    p_\gamma(z_t|c) \coloneqq p(z_t)^{1-\gamma}p(z_t|c)^\gamma.
    \label{eq: gamma-powered distribution}
\end{equation}
Diffusions and flows are grounded on solid modeling theory that ensures they excel at approximating individual scores
\begin{align}
    s(z_t) &\approx \nabla_{z_t}\log p(z_t),\label{eq: individual scores: unconditional}\\
    s(z_t|c) &\approx \nabla_{z_t}\log p(z_t|c),
    \label{eq: individual scores: conditional}
\end{align}
with which noise can be transformed into data. However, that theory has no approximation guarantees about the actual quantity of interest, namely the joint score
\begin{equation}
    \nabla_{z_t}\log p_\gamma(z_t|c).
    \label{eq: joint score}
\end{equation}
Most critically, the approximation
\begin{equation}
    \hat s_\gamma(z_t|c) = (1-\gamma)s(z_t) + \gamma s(z_t|c),
    \label{eq: cfg}
\end{equation}
does not guarantee that noise can be transformed into data.
Nevertheless, this is exactly what  CFG~\cite{CFG,Zheng_2023} does: it approximates \eqref{eq: joint score} with \eqref{eq: cfg}. In essence, CFG up-weights the probability of the samples where the conditional likelihood of an implicit classifier $p_i(c|z_t)$ is higher:
\begin{equation}
    \hat p_\gamma(z_t|c) \propto p(z_t|c)p_i(c|z_t)^{\gamma-1}.
    \label{eq: cfg_upweight}
\end{equation}

Continuous time diffusions and flows are trained~\cite{Kingma_Gao_2023} by abstracting the schedule $(\alpha_t, \sigma_t)$ via the log signal-to-noise ratio $\lambda_t\coloneqq\log(\alpha_t^2/\sigma_t^2)$ and training a parametric model $\epsilon_\theta(z_t, \lambda_t, c)$ by minimizing for each $(x, c)$
\begin{align*}
    &\mathcal{L}(x, c)\coloneqq\\
    &\mathbb{E}_{t\sim U(0,1), \epsilon\in\mathcal{N}(0, I)}\left\{w(\lambda_t)\cdot -\frac{d\lambda_t}{dt}\cdot\left\|\epsilon_\theta(z_t, \lambda_t, c)-\epsilon\right\|^2\right\}.
\end{align*}
Critically, gradient descent may yield $\theta^*$ such that
\begin{align}
    s(z_t|\emptyset) &\approx\epsilon_{\theta^*}(z_t, \lambda_t, \emptyset),\label{eq: diffusions and flows loss emptyset}\\
    s(z_t|c) &\approx\epsilon_{\theta^*}(z_t, \lambda_t, c).
    \label{eq: diffusions and flows loss c}
\end{align}

The continuous time formulation of diffusions and flows has recently been connected with the evidence lower bound (ELBO) formulation~\cite{Kingma_Gao_2023}.
Specifically, we leverage $v$-prediction with cosine schedule for which $w(\lambda_t) = e^{-\lambda_t/2}$ and
\begin{align*}
\mathcal{L}&=-\int_0^1\frac{dw(\lambda_t)}{dt}\mathbb{E}_{q(z_t|x,c)}\{\mathrm{ELBO}_t(z_t)\}dt + \mathrm{constant},\\
&\geq-\int_0^1\frac{dw(\lambda_t)}{dt}\mathbb{E}_{q(z_t|x,c)}\{\log p(z_t)\}dt + \mathrm{constant},
\end{align*}
where
\begin{equation*}
    \mathrm{ELBO}_t(z_t) \coloneqq \mathbb{E}_{q(z_{t+dt\ldots1}|z_t)}\left\{\log\frac{p(z_t,z_{t+dt\ldots1})}{q(z_{t+dt\ldots1}|z_t)}\right\}.
\end{equation*}

\subsubsection{CFG for Image Editing}

SOTA methods leverage CFG extended from single to multiple conditions. Given unedited image $c_I$ and instruction prompt $c_P$, the image editing task involves generating a new image which is the result of applying $c_P$ on $c_I$, posing it as an image generation model conditioned on both $(c_P, c_I)$. Notable work~\cite{brooks2022instructpix2pix} uses Eq.~\eqref{eq: alpha x + sigma epsilon} for implicit conditioning and provides $c_I$ as an additional input via concatenation with $z_t$. They use a pre-trained text-to-image model to provide a good initialization and incorporate $c_P$ as another input via cross-attention. As a result, they need to leverage CFG with respect to both conditionings to achieve plausible results where the denoising model is called three times at each iteration:
\begin{align}
&\hat \epsilon_\theta(\mathbf{z}_t, \lambda_t, \mathbf{c}_I, \mathbf{c}_P) \coloneqq \nonumber \\
& \epsilon_\theta(\mathbf{z}_t, \lambda_t, \varnothing, \varnothing ) \nonumber \\
& + s_I.\left[ \epsilon_\theta(\mathbf{z}_t, \lambda_t, \mathbf{c}_I, \varnothing ) - \epsilon_\theta(\mathbf{z}_t, \lambda_t, \varnothing, \varnothing)\right] \nonumber \\
& + s_P.\left[ \epsilon_\theta(\mathbf{z}_t, \lambda_t, \mathbf{c}_I, \mathbf{c}_P ) -  \epsilon_\theta(\mathbf{z}_t, \lambda_t, \mathbf{c}_I, \varnothing )\right],
\label{eq:cfg implicit}
\end{align}
 where $s_I$ and $s_P$ are the image and prompt guidance scales respectively. For a result that takes both unedited image and text prompt into account, we need all three passes by setting $s_I, s_P > 1$. We call this conditioning mechanism implicit because $(c_P, c_I)$ are implicitly embedded in $\mathbf{z}_t$. 

In the next section, we discuss explicit conditioning which outperforms CFG in terms of image generation quality and diversity but with just a single pass.

\subsection{Explicit conditioning} \label{sec:explicit_conditioning}
Instead of diffusing or flowing data into noise as in \eqref{eq: alpha x + sigma epsilon} where the conditioning term is implicitly embedded in $z_t$ via an additional input, we diffuse or flow data into a purpose-built distribution that explicitly incorporates the conditioning information:
\begin{equation}
    z_t \coloneqq \alpha_tx + \sigma_ty,\quad y\coloneqq g(\epsilon, f(\mathrm{condition})).
    \label{eq: alpha x + sigma y}
\end{equation}

To be consistent with the diffusion and flow model frameworks, we define $y$ to be sampled from Gaussian as follows:
\begin{align}
    y \sim \mathcal{N}(\mu_\psi(\mathbf{c}),\Sigma_\psi(\mathbf{c})*\mathbf{I}),
    \label{eq: y sim main}
\end{align}
where $\mu_\psi, \Sigma_\psi $ are functions that project a given conditioning term $\mathbf{c}$ into the same dimension as $\mathbf{x}$. Intuitively, instead of sampling from a normal Gaussian as done in traditional diffusions and flows, we sample from a Gaussian with mean and variance conditioned on functions of $\mathbf{c}$.

In the following, we discuss how the explicit modeling defined in Eq.~\eqref{eq: alpha x + sigma y} obliviates the need for CFG, and in particular reduces inference costs, without impacting generation quality or diversity.

\subsubsection{Interpretation in diffusion setting} \label{sec:theory_1}

From a diffusion point of view, let's assume we would apply CFG to Eq.~\eqref{eq: alpha x + sigma y}. Then according to Eq.~\eqref{eq: gamma-powered distribution}:
\begin{align*}
    \log p_\gamma(z_t|c) = (1-\gamma)\log p(z_t) + \gamma\log p(z_t|c).
\end{align*}
In particular, if $\log p(z_t|c)$ were to be constant as a function of $z_t$, then CFG would have no effect, since that term would disappear when taking the gradient to construct the score. {\em Could there be a modeling in which this is true?} Given Eq.~\eqref{eq: alpha x + sigma y}, one could argue that $p(z_t|c)$ `$=$' $p(z_t(c)|c)$ `$\approx$' $\mathrm{constant}$. This could happen if $t\approx 1$, \ie, early in the sampling where $\lambda_t \approx \lambda_{\min}$, or if the noise $\epsilon$ was approximately Dirac and $f$ and $g$ were regular enough with well-posed `inverse' $h$ such that $c = h(y, \epsilon)$. This constancy would imply $\nabla_{z_t}\log p(z_t|\mathrm{condition})\approx0$. This is insightful, but not always the case. Consider the counter-example: or $\delta\in(0,1)$, let $p_\delta:\mathbb{R}\rightarrow\mathbb{R}_0^+$ be a probability density function that is zero in $[-\infty,0]\cup[1,\infty]$ and is constant in $[\delta, 1-\delta]$ or equivalently an unnormalized density function $\tilde{p}_\delta(x)$ that is zero in $[-\infty,0]\cup[1,\infty]$ and is one in $[\delta, 1-\delta]$. If additionally $p_\delta$ and $\tilde{p}_\delta$ are smooth, then as $\delta\rightarrow0$ the gradient in $(0,\delta)\cup(1-\delta,1)$ must become arbitrarily large to transition between the constant values. Such an example can be constructed from classic mollifier functions:
\begin{figure}[t]
\centering
\begin{overpic}[width=0.5\textwidth]{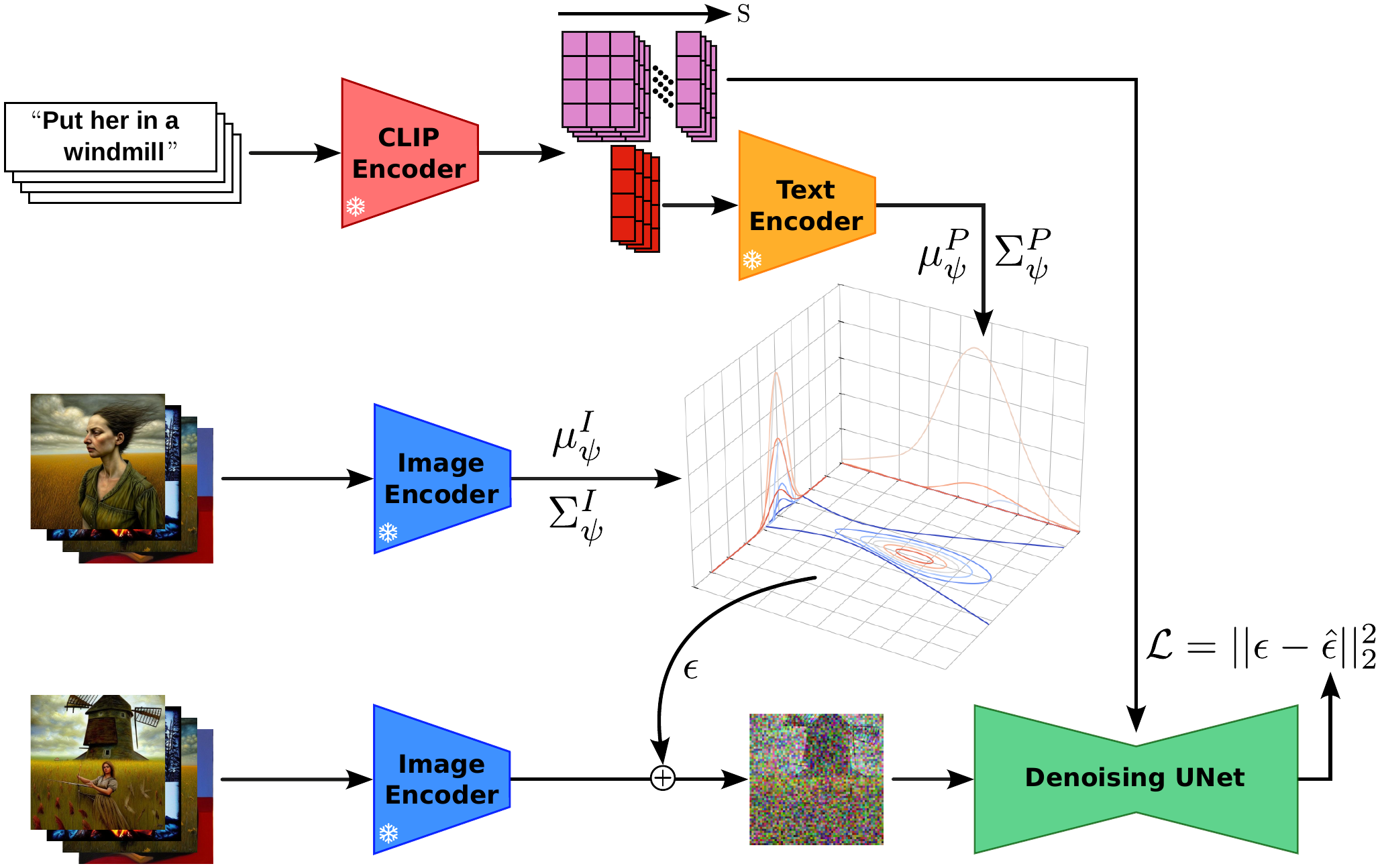}
  \put(1,88){Context}
  \put(5,35){Target}
\end{overpic}

\caption{\textbf{Explicit conditioning training:}\label{fig:method}{~ We obtain the mean and variance encoder of the context image  via the SD encoder, shown in blue. For the instruction prompt, we use our prompt VAE encoder, shown in yellow, that has the same latent dimension as the SD,~\ie, $4\times 64 \times 64$. It takes the pooled CLIP embeddings, shown in red, as input and maps them to the corresponding mean and variance. The context and prompt mean and variances are fused to form a Gaussian used for sampling in the diffusion process,  as in Eq.~\ref{eq: y sim context and prompt}}. We keep the full $77$ CLIP embeddings, shown in purple,  as input to the UNet for the sake of consistency with the internalization.}

\end{figure}

\begin{figure}[t]
\centering
\includegraphics[width=0.5\textwidth]{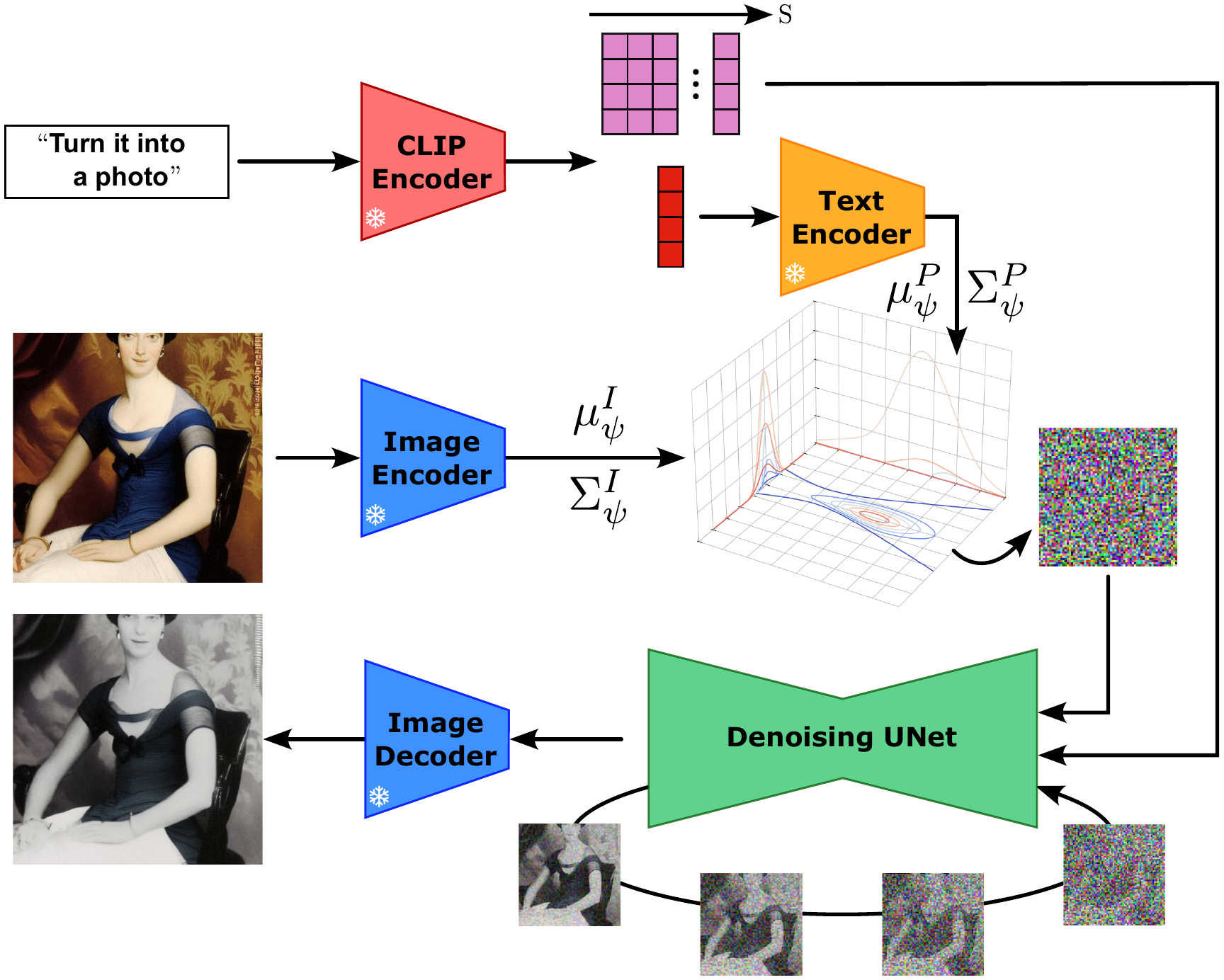}

\caption{\textbf{Inference:}\label{fig:inference} We call the denoising model recursively starting from a point sampled from a Gaussian that its mean and variance is a fusion of context image and instruction prompt.}

\vspace{-0.9cm}

\end{figure}

\begin{align*}
\tilde{p}_\delta(x)&\coloneqq
\begin{cases}
0 & x\leq0,\\
e^{\frac{-\delta^2}{\delta^2-(\delta-x)^2}+1} & x\in(0,\delta),\\
1 & x\in[\delta, 1-\delta],\\
e^{\frac{-\delta^2}{\delta^2-(x+\delta-1)^2}+1} & x\in(1-\delta,1),\\
0 & x\geq1,
\end{cases}\\
p_\delta(x)&\coloneqq\frac{\tilde{p}_\delta(x)}{\int\tilde{p}_\delta(z)dz}.
\end{align*}
Then, for any $\phi>0$, there exists $\delta>0$ such that $p_\delta(x)=1-\phi$ for $x\in(\delta, 1-\delta)$, hence $\frac{d}{dz}p_\delta(z)|_{z=x}=0$. However, for $\delta\ll1$, $\frac{d}{dz}p_\delta(z)|_{z=\frac{\delta}{2}}$ can be arbitrarily large. To demonstrate that these examples do not happen in diffusion and flow settings under EC in Eq.~\eqref{eq: alpha x + sigma y}, below we provide a different argument based on optimal transport.

\subsubsection{Flow optimal transport interpretation} \label{sec:theory_2}

Simply put, conditional generation in diffusion and flow models is done in the literature via additional inputs since the endpoint distributions do not contain that information. By putting forth EC in Eq.~\eqref{eq: alpha x + sigma y}, we open a new avenue of conditional modeling in which that information is present in one of the endpoint distributions. Given this, we ask: {\em Is there any need to pass that information additionally as inputs when approximating the score?} We show EC does not require additional inputs, in which case CFG is not needed by definition. There simply isn't any Eq.~\eqref{eq: individual scores: conditional}, only Eq.~\eqref{eq: individual scores: unconditional}. In other words, EC does flow-based optimal transport under the right distributions. For example: given unedited image $c_I$ and instruction prompt $c_P$ on the one hand, and edited image $x$ on the other hand, EC does flow-based optimal transport between $(c_I,c_P)$ and $x$.


\subsection{EC for image editing} \label{sec:ec}
In this section, we discuss in technical detail how we use EC for the image editing task. We build based on a widely used Stable Diffusion (SD)~\cite{Rombach_2022} framework and develop our method in two stages. First, we condition explicitly only on context, \ie, unedited image, and then extend the explicit conditioning for both context and the instruction prompt.
We show that incorporating context and instruction prompt results in guidance-free sampling, i.e., requires only $\times1$ pass.

\subsubsection{EC for Context}
The stable diffusion model framework operates in the latent space of a VAE that allows us to simply use it to obtain the mean and variance of the explicitly conditioned Gaussian on the context image, see Fig.~\ref{fig:method}
\begin{equation}
   \hat y \sim \mathcal{N}(\mu^I_\psi(c_I),\Sigma^I_\psi(c_I)*\mathbf{I}),
    \label{eq: y sim context}
\end{equation}
where $\mu^I_\psi,\Sigma^I_\psi$ are the pre-trained SD-VAE mean and variance estimators. Note that we use the same VAE to obtain edited image embeddings $x$, as a result, $y,x$ are of the same dimensions, ~i.e. $4\times 64\times 64$. Here, only $c_I$ is explicitly conditioned with the corresponding score function $\hat s(z_t, \lambda_t, c_P)$ that does not take $c_I$ as an additional input and $\hat z_t = \alpha_t x + \sigma_t \hat y$.

\if
CFG is required in this case only to guide the sampling with respect to the prompt, resulting in $\times 2$ sampling passes, see Fig.~\ref{fig:teaser} (c) and (d)
\begin{align}
& \hat \epsilon_\theta(\mathbf{z}_t, \lambda_t, \mathbf{c}_P) \coloneqq \nonumber \\
& \epsilon_\theta(\mathbf{z}_t, \lambda_t, \varnothing ) + s_P.\left[ \epsilon_\theta(\mathbf{z}_t, \lambda_t,\mathbf{c}_P ) -  \epsilon_\theta(\mathbf{z}_t, \lambda_t, \varnothing )\right].
\label{eq:cfg explicit context}
\end{align}
\fi

\subsubsection{EC for Context and Prompt}
To incorporate the prompt, we need to define a function $f$ in Eq.~\ref{eq: alpha x + sigma y} that takes $c_I, c_P$ as input and generates a tensor with the same dimension as $x$. For this purpose, we first train a tiny prompt VAE that takes pooled CLIP embeddings as input, and its latent space has the same dimension as $x$. During the image editing training via EC,  we use its frozen encoder to map pooled CLIP embeddings of the instruction prompt to a latent distribution of the same dimension as $x$, resulting in the following two functions:
$\mu^P_\psi(c_P):\mathbb{R}^{768} \rightarrow \mathbb{R}^{4\times 4\times 64}, \Sigma^P_\psi(c_P):\mathbb{R}^{768} \rightarrow \mathbb{R}^{4\times 64\times 64}$. We then sample $\tilde y$ from the following, see Fig.~\ref{fig:method}.
\begin{equation}
    \tilde y \sim \mathcal{N}(\mu^I_\psi(c_I) + \mu^P_\psi(c_P),(\Sigma^I_\psi(c_I) + \Sigma^P_\psi(c_P))*\mathbf{I}).
    \label{eq: y sim context and prompt}
\end{equation}
The corresponding score function,  $\tilde s(\tilde z_t, \lambda_t)$ where $\tilde z_t = \alpha_t x + \sigma_t \tilde y$,  is free from guidance for high-quality sampling,  see Fig.~\ref{fig:teaser},  and presumably independent of any auxiliary input.  However,  we initialize the model with a text-to-image model where full 77 CLIP tokens are fed to the score function via cross attention. We found that providing the model with the same CLIP tokens results in faster convergence and slightly better performance. We compare both cases w/ and w/o extra 77 CLIP tokens in our experimental section in Table.~\ref{tab:main}.

\paragraph{Prompt VAE architecture.} Prompt VAE is a tiny VAE with a latent space of dimension $4\times 64\times 64$. The encoder takes the $1\times768$ dimensional pooled CLIP embeddings of the instruction prompt as input and it has three layers. First, a linear layer projects the input from $768$ to $1024$ reshaped to $1\times 32\times 32$. Then the number of channels are increased to $16$ via a $1 \times 1$ Conv layer, resulting in a $16\times 32 \times 32$ tensor. We then use a parameter efficient spatial upsampling operation of PixelShuffle~\cite{pixelshuffle} to obtain a $4 \times 64 \times 64$ tensor. The mean and variance are obtained via two separate $1 \times 1$ Conv layers. The decoder mirrors the encoder. We train the encoder and decoder with standard VAE losses, ~\ie,~$l_2$ reconstruction, and KL losses. A LeakyReLU and an instance normalization layer follow all Conv and projection layers. The rationale behind our design choices here is to first, keep the amount of changes to the original signal minimal and second, transform it to the desired dimension that has a Gaussian distribution.

\paragraph{Sampling.}
During the inference, for a given $c_I, c_P$, we first obtain their corresponding mean variances via image and prompt encoders. We then sample a starting point from Eq.~\ref{eq: y sim context and prompt} and perform $10$  DDIM steps, see Fig.~\ref{fig:inference}.

\section{Experiments} \label{sec:experiments}

In this section, we evaluate our proposed Explicit Conditioning (EC) method for the image editing task and compare it with Classifier Free Guidance (CFG). Moreover, we introduce a new metric \emph{Directional Visual Similarity} (DVS), as an alternative to directional CLIP similarity which is more aligned with the training objective function. Our proposed approach outperforms quantitatively and qualitatively ~\cite{brooks2022instructpix2pix} which uses CFG during the inference, while being more robust and computationally efficient at the same time. We regenerate the results of ~\cite{brooks2022instructpix2pix} and use 10 sampling steps for all methods throughout the experiments.

\subsection{Experimental setup.}
We rely on Stable Diffusion SDv1.5 for its affordable computational cost and performance trade-off. The model is fine-tuned for image editing on the Instruct-pix2pix dataset~\cite{brooks2022instructpix2pix}, and evaluated on the full test set. 

\subsection{Evaluation metrics.}

We rely on two metrics to assess the quality of the edits: (I) the widely used Directional CLIP Similarity~\cite{gal2021stylegannada} (DCS), and (ii) our proposed Directional Visual Similarities (DVS) that we discuss in the following. Intuitively, DCS measures the alignment between the performed edit -- from the generated  image $o$ to the input one $i$ -- and the change in captions -- of the input image $i$ and the target image $e$. This alignment is measured in the latent space through CLIP encoders:
\begin{align*}
DCS = S_C ( [\mathrm{E_V}(i) -\mathrm{E_V}(o)], [\mathrm{E_T}(c_i) - \mathrm{E_T}(c_e)]  ) ,
\end{align*}
where $S_C$ is the cosine similarity and  $\mathrm{E_V, E_T}$ are CLIP's vision and text encoders respectively. This metric uses text embeddings because it takes into account that the training dataset is synthetic where the input and edited target images are generated synthetically via text-to-images using their corresponding captions. As a result, the target image truthfulness depends on the image synthesizer's behavior, making it a pseudo-ground-truth. Note that the captions are not presented during the training. Therefore, this metric suffers from the misalignment between training and test distributions injected by the image generator. Moreover, the deviation between the CLIP vision and text encoders imposes another bias.
To address these issues, we introduce \emph{Directional Visual Similarity} which is more aligned with the training objective function and robust to the deviations between the CLIP vision and text encoders:
\begin{align*}
DVS = S_C ( [\mathrm{E_V}(i) -\mathrm{E_V}(o)], [\mathrm{E_V}(i) -\mathrm{E_V}(e)]  ) .
\end{align*}
Where $S_C$ is the cosine similarity and $E_V$ is a vision encoder. We use CLIP vision encoder to be consistent with DCS. This metric measures the alignment of the transformation context-to-generated image -- $i$ to $o$ --, and the transformation context-to-target image -- $i$ to $e$. Therefore, DVS is more aligned with the training objective where the task is to generate the target image regardless of how they have been generated. As it is shown in Table ~\ref{tab:main}, DCS and DVS are correlated, however, DCS is more sensitive to the true output that might not be depicted in the pseudo-gt. On the other hand, DVS is more consistent with our qualitative assessment.

\begin{table}[t]
\centering
\resizebox{\columnwidth}{!}{ 
\begin{tabular}{c c c c  }
Method & Denoisng passes $\downarrow$ & DCS $\uparrow$ & DVS $\uparrow$ \\
\hline
CFG & $\times 1$ & $0.114$ & $0.390$ \\
CFG & $\times 3$ & $0.177$ & $0.448$ \\
\midrule
EC(ours) w/ extra tokens & $\times 1$ & $\textbf{0.191}$ & $0.576$ \\
EC(ours) w/o extra tokens & $\times 1$ & $0.178$ & $\textbf{0.578}$ \\
\bottomrule
\end{tabular} }
\caption{\textbf{Quantitative evaluation.} We compare our explicit conditioning method with CFG using the widely used directional clip similarity (DCS) metric as well as our newly proposed directional visual similarity (DVS). We consider two cases for CFG: $\times 1$ pass, \ie,~ $s_I=1.0, s_P=1.0$, and $\times 3$ passes $s_I=1.6, s_P=7.5$. CFG's performance significantly improves with $\times 3$ more computational cost, however, it still underperforms ours which requires only  $\times 1$ pass.} 
\label{tab:main}
\vspace{-0.5cm}
\end{table}

\subsection{Quantitative evaluation.}
We consider the implicit conditioning mechanism as a baseline that heavily relies on CFG via Eq.~\ref{eq:cfg implicit}. We use the pre-trained model by ~\cite{brooks2022instructpix2pix} and to see the impact of CFG, we evaluate $\times 1$ pass, ~\ie~ $s_I=1.0, s_P=1.0$ and $\times 3$ passes, ~\ie~ $s_I=1.6, s_P=7.5$\footnote {Note that it is infeasible to find an optimal set of guidance scales for the entire dataset as it is sample dependent. We choose these values according to the evaluations in ~\cite{brooks2022instructpix2pix}.}. We compare with our explicitly conditioned model on both context image and instruction prompt via Eq.~\ref{eq: y sim context and prompt}, which is guidance-free and requires only $\times 1$ pass. Table~\ref{tab:main} summarizes the results.  We compare two cases of our method w/ and w/o the 77 CLIP tokens as additional input to the UNet. We found that providing the UNet with the 77 extra CLIP tokens results in a faster convergence and better DCS score, which could be due to consistency with the text-to-image initialization model. Our proposed EC method outperforms CFG in both cases in terms of DCS and DVS metrics while being $\times 3$ faster.

\subsection{Qualitative evaluation.}
\begin{figure}[t!]
\begin{minipage}{\columnwidth}
\centering
\textbf{prompt:} Turn the lake into a river.
\end{minipage}

\begin{minipage}{0.3\columnwidth}
\includegraphics[width=\columnwidth]{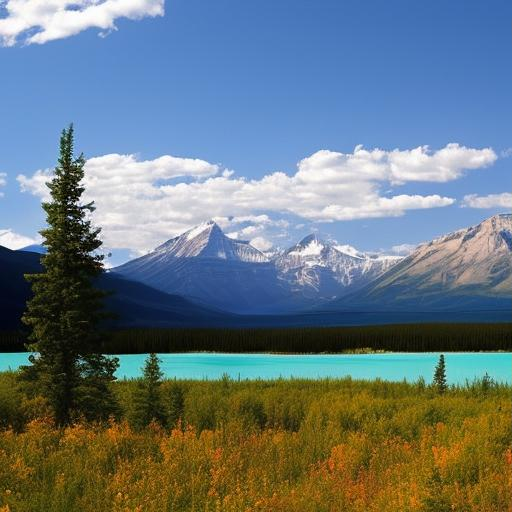}
\end{minipage}
\begin{minipage}{0.3\columnwidth}
\includegraphics[width=\columnwidth]{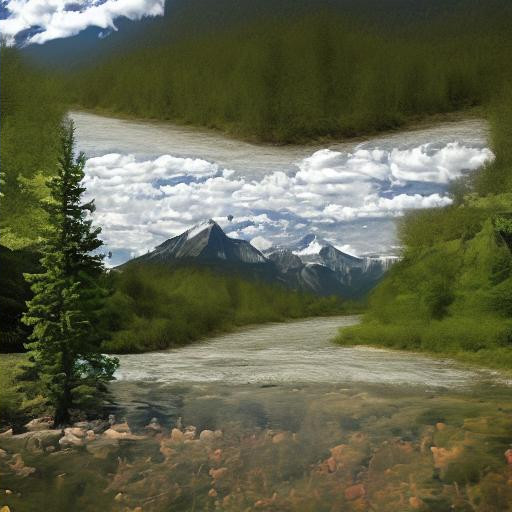}
\end{minipage}
\begin{minipage}{0.3\columnwidth}
\includegraphics[width=\columnwidth]{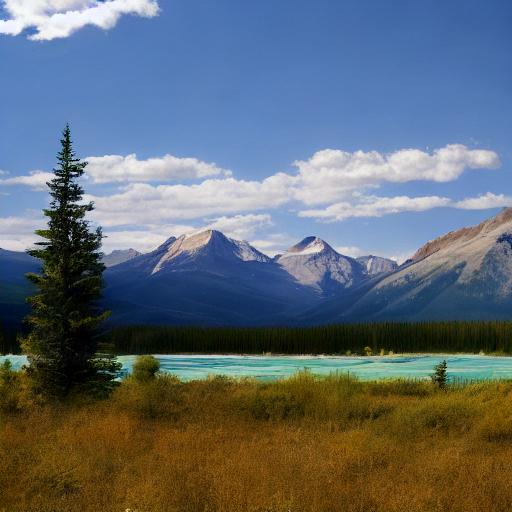}
\end{minipage}

\begin{minipage}{0.3\columnwidth}
\centering
(a) context
\end{minipage}
\begin{minipage}{0.3\columnwidth}
\centering
(b) CFG 
\end{minipage}
\begin{minipage}{0.3\columnwidth}
\centering
(c) EC
\end{minipage}

\begin{minipage}{\columnwidth}
\centering
\textbf{prompt:} It is a full moon.
\end{minipage}

\begin{minipage}{0.3\columnwidth}
\includegraphics[width=\columnwidth]{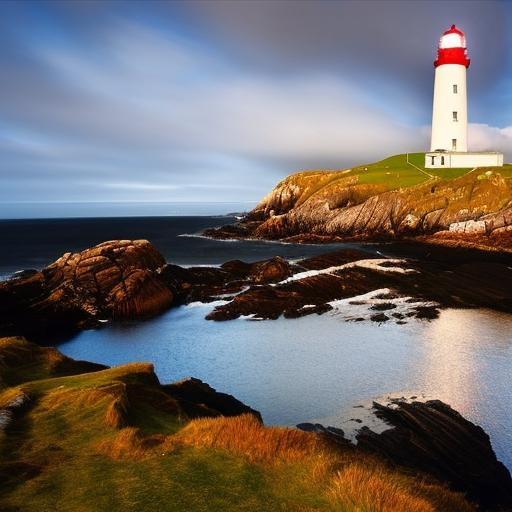}
\end{minipage}
\begin{minipage}{0.3\columnwidth}
\includegraphics[width=\columnwidth]{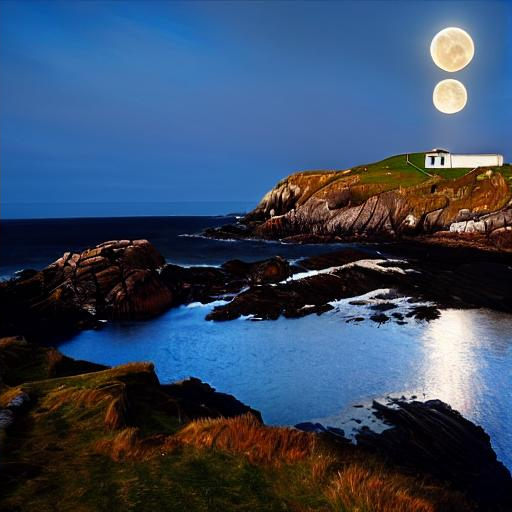}
\end{minipage}
\begin{minipage}{0.3\columnwidth}
\includegraphics[width=\columnwidth]{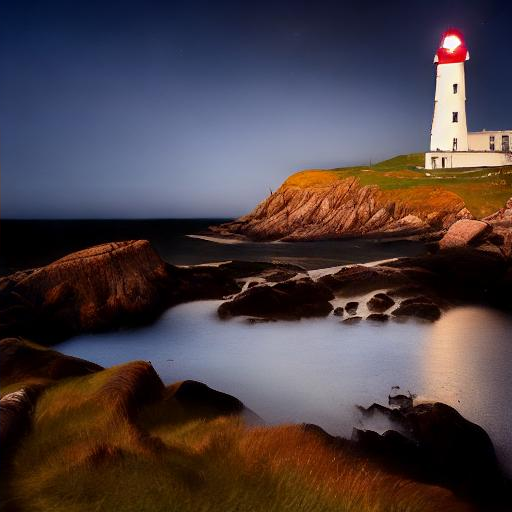}
\end{minipage}

\begin{minipage}{0.3\columnwidth}
\centering
(a) context
\end{minipage}
\begin{minipage}{0.3\columnwidth}
\centering
(b) CFG 
\end{minipage}
\begin{minipage}{0.3\columnwidth}
\centering
(c) EC
\end{minipage}

\caption{Qualitative evaluations, failure cases. (a) input context image, (b) CFG $\times 3$ passes, \ie~$s_I=1.6, s_P=7.5$. (c) our proposed $\times 1$ pass explicit conditioning.}
\label{fig:failures}
\vspace{-0.4cm}
\end{figure}

\begin{figure*}[ht!]
\begin{minipage}{\textwidth}
\centering
\textbf{prompt:} Make the characte wear a yellow dress.
\end{minipage}

\begin{minipage}{0.245\textwidth}
\includegraphics[width=\textwidth]{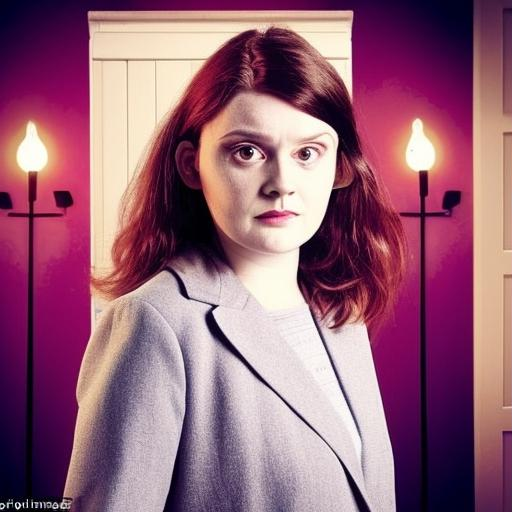}
\end{minipage}
\begin{minipage}{0.245\textwidth}
\includegraphics[width=\textwidth]{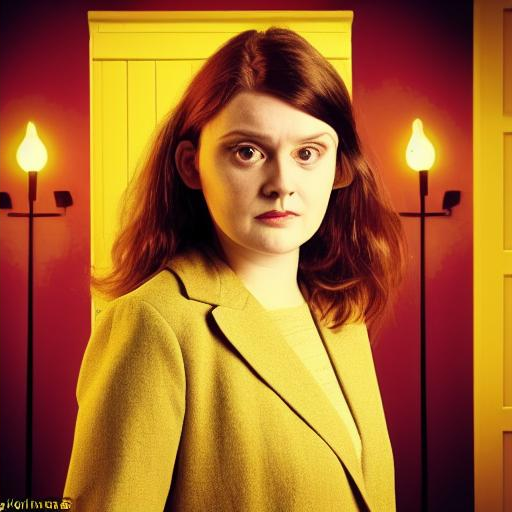}
\end{minipage}
\begin{minipage}{0.245\textwidth}
\includegraphics[width=\textwidth]{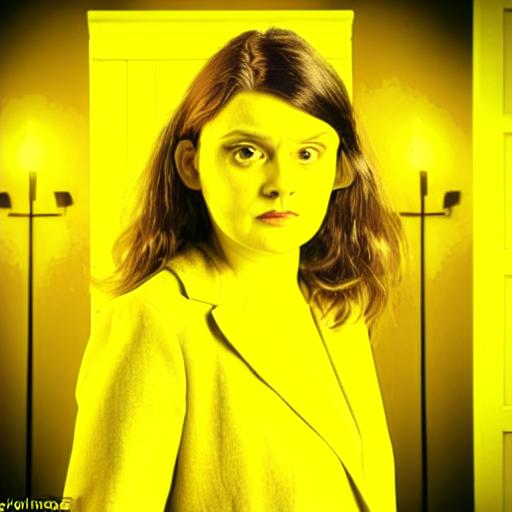}
\end{minipage}
\begin{minipage}{0.245\textwidth}
\includegraphics[width=\textwidth]{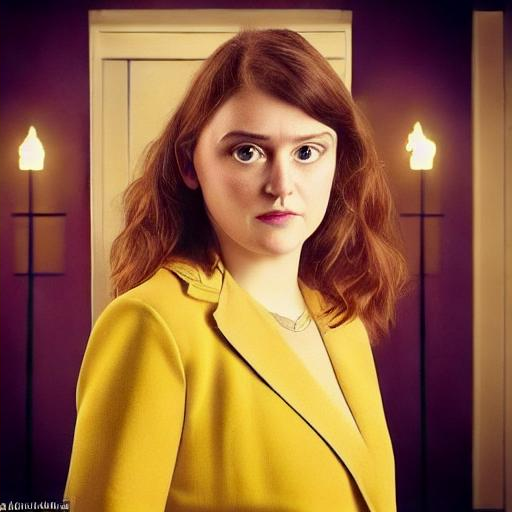}
\end{minipage}

\begin{minipage}{0.245\textwidth}
\centering
(a) context
\end{minipage}
\begin{minipage}{0.245\textwidth}
\centering
(b) CFG $\times 1$ pass
\end{minipage}
\begin{minipage}{0.245\textwidth}
\centering
(c) CFG $\times 3$ passes
\end{minipage}
\begin{minipage}{0.245\textwidth}
\centering
(d) EC $\times 1$ pass
\end{minipage}

\begin{minipage}{\textwidth}
\centering
\textbf{prompt:} Make the canyon look like a giant cave.
\end{minipage}

\begin{minipage}{0.245\textwidth}
\includegraphics[width=\textwidth]{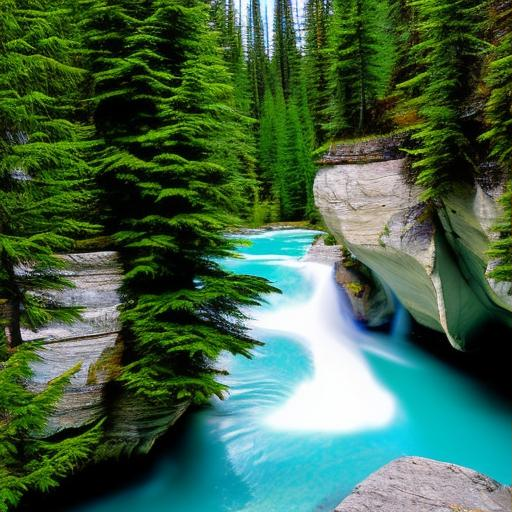}
\end{minipage}
\begin{minipage}{0.245\textwidth}
\includegraphics[width=\textwidth]{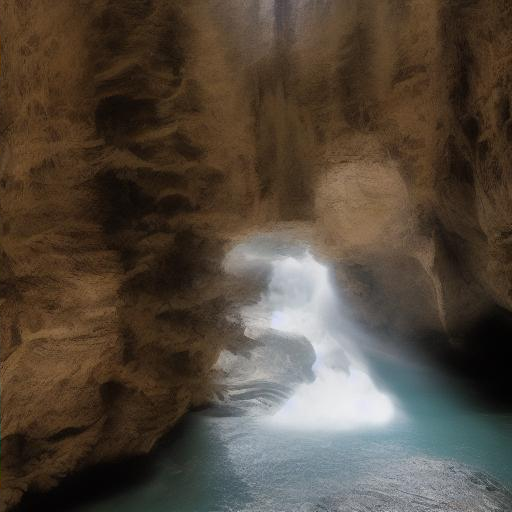}
\end{minipage}
\begin{minipage}{0.245\textwidth}
\includegraphics[width=\textwidth]{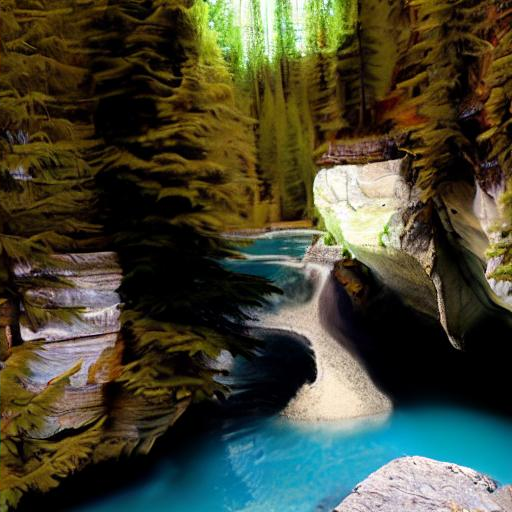}
\end{minipage}
\begin{minipage}{0.245\textwidth}
\includegraphics[width=\textwidth]{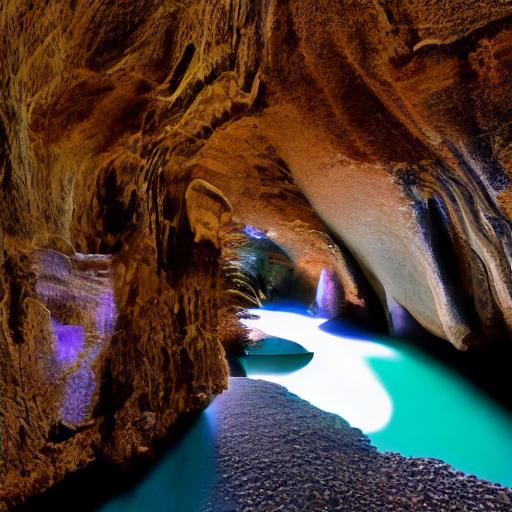}
\end{minipage}

\begin{minipage}{0.245\textwidth}
\centering
(a) context
\end{minipage}
\begin{minipage}{0.245\textwidth}
\centering
(b) CFG $\times 1$ pass
\end{minipage}
\begin{minipage}{0.245\textwidth}
\centering
(c) CFG $\times 3$ passes
\end{minipage}
\begin{minipage}{0.245\textwidth}
\centering
(d) EC $\times 1$ pass
\end{minipage}

\begin{minipage}{\textwidth}
\centering
\textbf{prompt:} Make it a winter scene.
\end{minipage}

\begin{minipage}{0.245\textwidth}
\includegraphics[width=\textwidth]{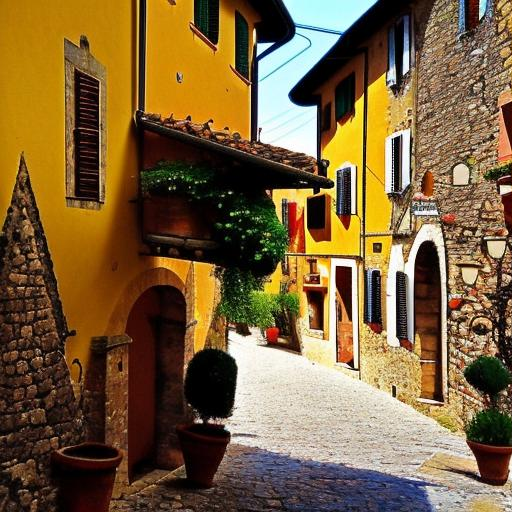}
\end{minipage}
\begin{minipage}{0.245\textwidth}
\includegraphics[width=\textwidth]{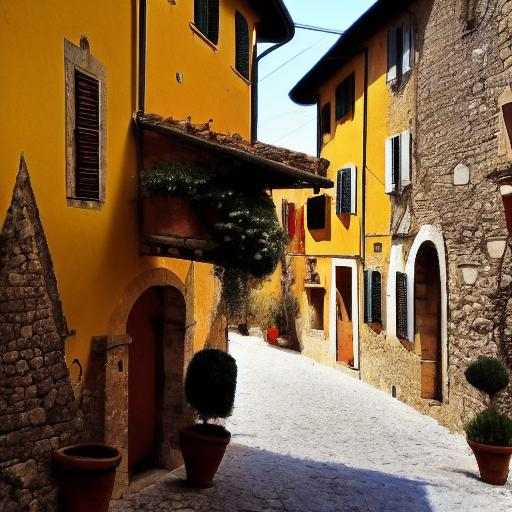}
\end{minipage}
\begin{minipage}{0.245\textwidth}
\includegraphics[width=\textwidth]{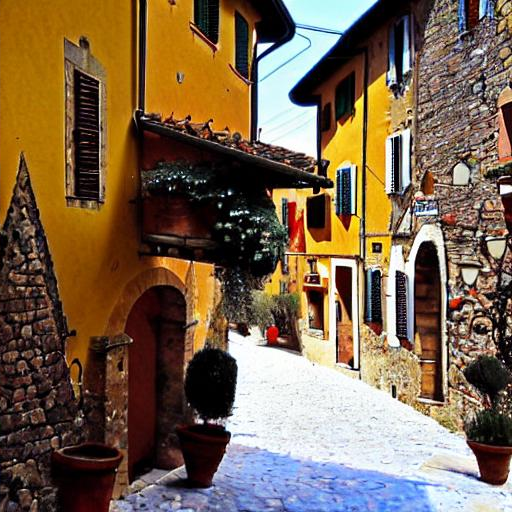}
\end{minipage}
\begin{minipage}{0.245\textwidth}
\includegraphics[width=\textwidth]{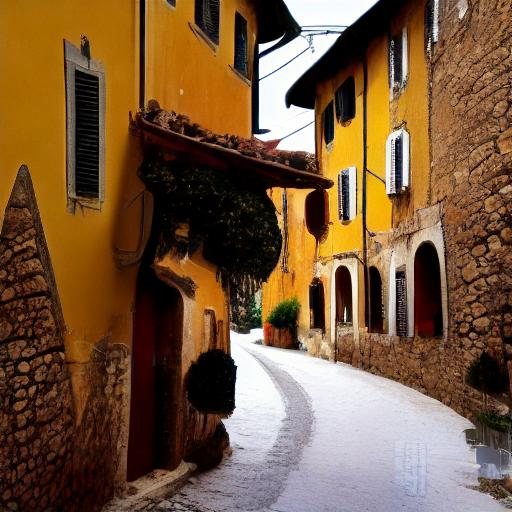}
\end{minipage}

\begin{minipage}{0.245\textwidth}
\centering
(a) context
\end{minipage}
\begin{minipage}{0.245\textwidth}
\centering
(b) CFG $\times 1$ pass
\end{minipage}
\begin{minipage}{0.245\textwidth}
\centering
(c) CFG $\times 3$ passes
\end{minipage}
\begin{minipage}{0.245\textwidth}
\centering
(d) EC $\times 1$ pass
\end{minipage}

\begin{minipage}{\textwidth}
\centering
\textbf{prompt:} Make it a hotel.

\end{minipage}

\begin{minipage}{0.245\textwidth}
\includegraphics[width=\textwidth]{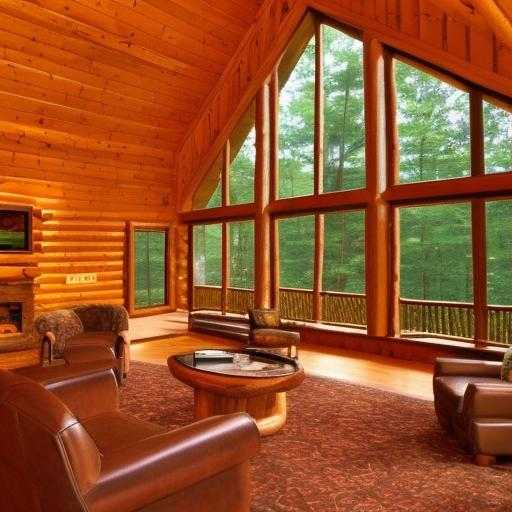}
\end{minipage}
\begin{minipage}{0.245\textwidth}
\includegraphics[width=\textwidth]{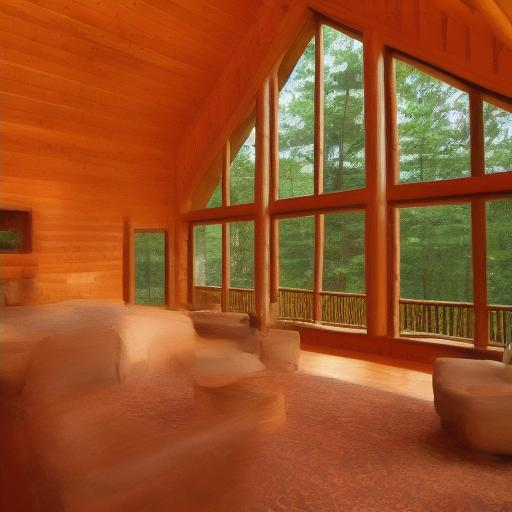}
\end{minipage}
\begin{minipage}{0.245\textwidth}
\includegraphics[width=\textwidth]{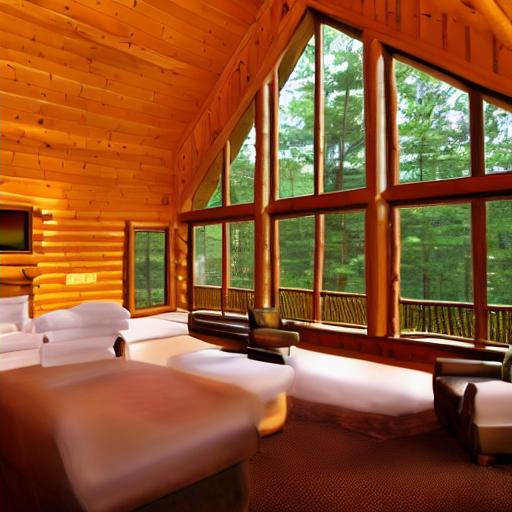}
\end{minipage}
\begin{minipage}{0.245\textwidth}
\includegraphics[width=\textwidth]{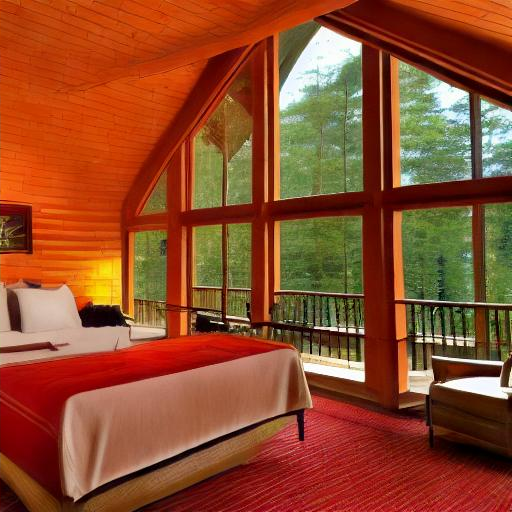}
\end{minipage}

\begin{minipage}{0.245\textwidth}
\centering
(a) context
\end{minipage}
\begin{minipage}{0.245\textwidth}
\centering
(b) CFG $\times 1$ pass
\end{minipage}
\begin{minipage}{0.245\textwidth}
\centering
(c) CFG $\times 3$ passes
\end{minipage}
\begin{minipage}{0.245\textwidth}
\centering
(d) EC $\times 1$ pass
\end{minipage}
\caption{{\bf Qualitative evaluations.} (a) input context image, (b) CFG $\times 1$ pass, \ie $s_I=1.0, s_P=1.0$ (c) CFG $\times 3$ passes, \ie~$s_I=1.6, s_P=7.5$. (d) our proposed $\times 1$ pass explicit conditioning. Our method outperforms CFG while being $\times 3$ faster.
\newline
\newline
\newline}
\label{fig:qualitative}

\end{figure*}

Fig.~\ref{fig:qualitative} compares qualitatively our single-pass explicit conditioning model with CFG. We consider two cases for CFG with $\times 1$ pass, \ie,~ $s_P=1, s_I=1$ in Eq.~\ref{eq:cfg implicit} which is computationally similar to us, and $\times 3$ passes, \ie,~$s_P=1.6, s_I=7.5$. Fig.~\ref{fig:qualitative}  supports the quantitative results in Tab.~\ref{tab:main} where our proposed explicit conditioning significantly outperforms CFG while being $\times 3$ faster. Moreover, Fig.~\ref{fig:failures} shows some failure cases of our method. 

\section{Limitations} \label{sec:limitations}
Although our method is more efficient and robust compared with CFG, however, the impact of the input context can be only controlled by the instruction prompt. The guidance scale in CFG can be considered as a tool to control the impact of the context image. While this can be seen as an advantage, it makes CFG ill-conditioned to find the optimal scale and computationally inefficient. As we show in our experiments, explicit conditioning is superior in following the given instructions for image editing, as a result, our proposed method benefits more significantly from a higher-quality dataset where more diverse and detailed instructions are provided with instructions such as "make it more/less look like the input".

\section{Conclusion} \label{sec:conclusion}
We proposed explicit conditioning for image editing. Our main idea involves diffusing the data distribution into a purposefully designed
distribution made of noise and condition. We theoretically discussed how our proposed method results in a guidance-free sampling mechanism and showed via qualitative and quantitative evaluations that we outperform popular classifier-free guidance sampling while being $\times 3$ faster. We believe our proposed method opens a new direction in the rapidly progressing area of conditional generative models and can be potentially extended to other generative modeling tasks and frameworks such as flow models.


\bibliographystyle{icml2025}
\bibliography{main}




\end{document}